\documentclass[a4paper,twoside]{article}

\usepackage{epsfig}
\usepackage{subcaption}
\usepackage{calc}
\usepackage{amssymb}
\usepackage{amstext}
\usepackage{amsmath}
\usepackage{amsthm}
\usepackage{multicol}
\usepackage{pslatex}
\usepackage{apalike}
\usepackage{hyperref}
\usepackage{SCITEPRESS}     

\begin{document}

\title{PolarNet: Accelerated Deep Open Space Segmentation
	\\Using Automotive Radar in Polar Domain}

\author{\authorname{Farzan Erlik Nowruzi\sup{1,2},
		Dhanvin Kolhatkar\sup{2},
		Prince Kapoor\sup{2},
		Elnaz Jahani Heravi\sup{2},
		Fahed Al Hassanat\sup{1,2},
		Robert Laganiere\sup{1,2},
		Julien Rebut\sup{3},
		Waqas Malik\sup{3}}
	\affiliation{\sup{1}School of Electrical Engineering and Computer Sciences, University of Ottawa, Canada}
	\affiliation{\sup{2}Sensorcortek Inc, Canada, \sup{3}Valeo, France}
	\email{fnowr010@uottawa.ca,\{dhanvin, prince, elena\}@sensorcortek.ai,\\ falha023@uottawa.ca, laganier@eecs.uottawa.ca, \{julien.rebut, waqas.malik\}@valeo.com}
}


\abstract{Camera and Lidar processing have been revolutionized with the rapid development of deep learning model architectures. Automotive radar is one of the crucial elements of automated driver assistance and autonomous driving systems. Radar still relies on traditional signal processing techniques, unlike camera and Lidar based methods. We believe this is the missing link to achieve the most robust perception system. Identifying drivable space and occupied space is the first step in any autonomous decision making task. Occupancy grid map representation of the environment is often used for this purpose. In this paper, we propose PolarNet, a deep neural model to process radar information in polar domain for open space segmentation. We explore various input-output representations. Our experiments show that PolarNet is a effective way to process radar data that achieves state-of-the-art performance and processing speeds while maintaining a compact size.}

\onecolumn \maketitle \normalsize \setcounter{footnote}{0} \vfill

\section{\uppercase{Introduction}}
\label{introduction}

Autonomous driving can be approached from various sensor modalities such as camera, radar, sonar, or lidar.

Each sensor provides a piece of valuable information. It is shown that the fusion of multiple sensors is required to cover all of the environmental variations and achieve fully autonomous driving. Despite this observation, each individual sensor needs to be pushed to the edge of its capabilities to reduce the complexity of the fusion systems.

Radar sensors have been around in the automotive industry for a few decades already. The first systems were mostly used in the premium car segment for comfort applications, such as Adaptive Cruise Control (ACC). With the continuous improvement of radar technology, recent radar sensors are also used in safety applications, such as Autonomous Emergency Braking (AEB). 


Furthermore, radars are not affected by poor lightning and fog, unlike camera and Lidar respectively. The new versions of radar will significantly increase the resolution of the observations, therefore enabling an even wider variety of applications.

Ultra-sonic sensors are very similar to radar, with a focus on close range applications only, such as automated parking detection. In this paper, we propose a model to equip radar with the ability to replace ultra-sonic sensors. In this way, a single sensor can run in multiple modes to provide better value and reduce system complexity.

In recent years, deep learning models have achieved state-of-the-art performance on various applications \cite{InceptionResNet}\cite{fasterrcnn}\cite{yolov3}\cite{retinanet}\cite{2017_deeplabv3}\cite{maskrcnn}\cite{pointnet}. However, these models are rarely used with automotive radar; traditional signal processing methods such as Constant False Alarm Rate (CFAR) \cite{cfar1986}\cite{cfar1990} are commonly used to process radar data. Radar is an alternative depth sensor that presents a less expensive solution than Lidar at the cost of quality of depth. Recent advances in radar technology introduced High Definition radars that address the issues with the quality of the depth data. The lack of publicly available radar datasets could be seen as the primary reason for the smaller amount of literature in this field. This issue is partially addressed by novel datasets with radar observations ~\cite{oxfordRadar2019}\cite{sc2020icmim} that have recently been introduced to enable the scientific community to expand the boundaries of knowledge in this field.

In this paper, we introduce a novel deep learning approach that takes benefit of the polar representation of the radar observations and performs open space segmentation in parking lot scenarios. 
In the past, these applications relied on traditional signal processing techniques such as Constant False Alarm Rate (CFAR) that employs hand crafted filters to detect points on a potential object. 
The area can then be segmented into occupied and available spaces based on these points.
We explore using learnable filters, which have shown improvements over hand-crafted ones in many computer vision fields.
Our model relies on a series of \textit{1D} and \textit{2D} convolutions that reaches state-of-the-art segmentation performance in an end-to-end manner. To address the requirements of the automotive industry, we ensured that the model is compact and fast to run on embedded platforms. Our contributions are listed as follows: 

\begin{enumerate}
	
	\item A deep model, PolarNet, that takes radar frames in polar coordinate and generates an open-space segmentation mask in a parking-lot scenario
	
	
	\item Evaluation of various models and loss functions for this task
	
	\item Detailed comparison of PolarNet to the state-of-the-art in terms of performance, speed, and size.
	
\end{enumerate}

The literature review of deep segmentation model architectures, along with radar applications, are discussed in Section~\ref{litrev}. A brief description of radar data can be found in Section~\ref{dataset}. The compared model architectures are described in Section~\ref{model}. In Section~\ref{exps}, various experiments, including the effect on model performance of using different input data representations, segmentation models, and loss functions are evaluated. Furthermore, the computational complexity requirements of each model are discussed in detail.

\section{\uppercase{Literature Review}}
\label{litrev}

Many research works are proposed to address the challenge of labeling individual pixels in images for semantic segmentation. There are two category of methods in this subject. One uses an \textit{encoder-decoder} architecture, and the other uses specialized convolutions to avoid decimating input map size. The complexity of the former approach is highly dependent on the models used for each component, while the latter suffers from large memory requirements caused by maintaining the large feature maps which help in generating fine segmentation masks.


\cite{chen2014semantic} first proposed the idea of using a fully connected conditional random field as a decoder at the end of a deep convolutional model to extract quality segmentation masks.

\textit{Fully Convolutional Network} (FCN)~\cite{FCN} uses an encoder network, made up of convolutional layers only, to extract intermediate feature maps of the inputs. By employing skip-connections, upsampling, and transposed convolution operations on the decoder side, an output mask of the desired size is generated.

Following up with the idea of \cite{chen2014semantic}, \textit{DeconvNet} is introduced in \cite{noh2015learning}, and utilizes a more specialized decoder architecture that consists of a series of decoupled unpooling and convolution layers.

\cite{segnet} proposes a similar architecture to DeconvNet, but with greatly reduced complexity and compares variations of the FCN, DeconvNet and SegNet models.

The \textit{U-Net}\cite{unet}  architecture iterates on the FCN model by using deeper decoder feature maps and extensive data augmentation. The \textit{U-Net++}~\cite{unet++} architecture is a more general version of U-Net which significantly increases the number and the complexity of the skip connections between the encoder and the decoder.

\cite{2016_deeplabv2} pioneered the use of atrous convolutions for segmentation to avoid subsampling of the input data, and instead enlarge the field of view of the convolution kernel without using any pooling layers. Furthermore, multiple parallel atrous convolutions are utilized to segment objects at different scales.

\cite{pspnet} introduced the idea of pyramid pooling to collect global and local information. \cite{InceptionResNet} is used with \cite{2016_deeplabv2} to build mid-level feature maps. By using parallel pooling layers, coarse to fine information is generated that is later used to extract the final segmentation result. \cite{2018_deeplabv3+} merged both categories of methods by using \cite{2016_deeplabv2} as the encoder network and using a small decoder network to achieve better performance.


To accelerate scene segmentation, \cite{badino2009stixel} considers the space in front of a vehicle free unless there is a vertical obstacle present. This resulted in the introduction of Stixel as a rectangular block on the image that identifies vertical surfaces. Hence, a compressed representation of the environment is achieved. \cite{schneider2016semantic} adds the semantic labels to each stixel. Inspired by these ideas, Stixelnet \cite{levi2015stixelnet} segments an image for open space using stixel-like rectangular regions. Instead of a segmentation model, it relies on a classification network that predicts the junction point for ground and the obstacle. Later, a conditional random field is used to smooth the jittery predictions from the column-wise network. In this case, each stixel corresponds to a specific angle interval from the camera point of view. As we explain later, Polar representation of the radar is well suited for this family of architectures.

The majority of aforementioned methods rely on camera based inputs. It is common to use a pre-trained back-bone to build a new model as training a completely new architecture is still a challenging problem. However, in the case of radar input data, using a pre-trained model from other domains is not a promising approach. This is due to the fact that there are various radar representations, and that most radar processing currently uses traditional techniques \cite{cfar1986}\cite{cfar1990}. Sless et al.~\cite{sless2019road} proposes an encoder-decoder architecture with a Bird's Eye View (BEV) input and a 3-class output: occupied, unoccupied or unobservable.

Bauer et al.~\cite{2019_deepISM} proposed a similar U-Net architecture for occupancy prediction. They formulate the classification problem as both a three class problem, as in \cite{sless2019road}, and as a four class problem. The four-class approach uses an \textit{unknown} class to show the certainty of the predictions.

Another issue in the radar domain is the low number of publicly available datasets that contain some form of radar data. The NuScenes dataset \cite{nuscenes2019} is a large-scale dataset developed for autonomous driving. However, the radar data is provided after processing with traditional models, rather than providing the raw signal data.

The \textit{Oxford Radar RobotCar} dataset~\cite{oxfordRadar2019} is available for scene understanding analysis. The radar used for data collection offers much finer resolution and higher range than typical automotive radars. It provides a $360$ degree azimuth-range representation of the received power reflection. It is worth noting that this is 2 dimensional representation. Raw radar data is not available in this dataset, but the radar modality provided by the authors is less processed than that of the NuScenes dataset. The usage of specialized hardware uncommon in the automotive domain due to its physical characteristics is a major disadvantage for some applications. Most recently, \cite{sc2020icmim} introduced a novel dataset for open space segmentation in automotive parking scenarios. The dataset includes raw radar echos collected by an automotive grade radar along with the tool chain to extract various representations. They proposed a variation of FCN targeted for embedded platform deployment. Their results are evaluated against well-known deep segmentation models. We rely on this dataset to benchmark our work.

\section{\uppercase{Radar Data}}
\label{dataset}

To propose a new model, we first need to understand the data that will be used. Raw radar data consists of a series of echos received from each transmitter-receiver combination. Each signal consists of multiple chirps and samples. This results in a four dimensional matrix of Samples, Chirps, Transmitters, and Receivers. It is common that the last two dimensions are concatenated and stacked as one dimension resulting in a three dimensional representation of Samples, Chirps, and Antennas (SCA). SCA is a raw echo representation. Although it includes all the information, its visualization does not provide much insight into the data.

To address this, Fast Fourier Transforms (FFT) are applied along each dimension to achieve the Range, Doppler, and Azimuth (RDA) representation. The distance associated with an observation is represented in the \textit{Range} dimension, the velocity is represented in the \textit{Doppler} dimension, and the angle of arrival of the signal is represented in the \textit{Azimuth} dimension. Note that prior to applying the FFT along the Antenna dimension, zero-padding is applied along the last dimension to achieve the target angle resolution.

\begin{equation}
\begin{aligned}[c]
RDA = \mathfrak{fft}_{A}\bigg( \mathtt{zero\_pad_{A}}\Big( \mathfrak{fft}_{C}\big(\mathfrak{fft}_{S}(SCA)\big) \Big) \bigg)
\end{aligned}
\end{equation}


RDA representation is a 3D tensor that, depending on the view point, shows multiple representations. For the case of open space segmentation, the most suitable representation is the RAD view that generates a polar Range-Azimuth map for each Doppler bin.

\begin{figure}
	\begin{center}
		\includegraphics[width=0.99\linewidth]{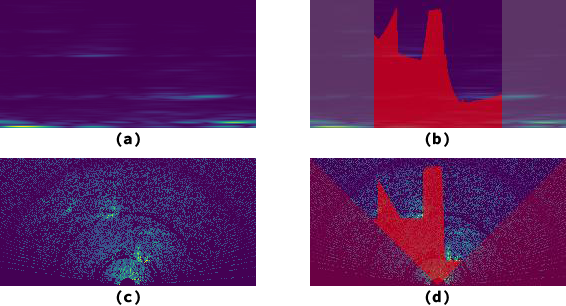}
	\end{center}
	\caption{Samples from the dataset. (a) Range-Azimuth representation. (b) Ground truth annotation on RA. (c) RA in Cartesian form. (d) Ground-truth projected on Cartesian map. The angle interval between $-45$ and $+45$ is selected for training.}
	\label{fig:dataset}	
\end{figure}

Taking log and summing along the Doppler dimension results in Range-Azimuth (RA) representation. RA is a simple 2D BEV of radar data in polar coordinates.

\begin{equation}
R_{i}A_{k} = \sum^{D}_{j=0} \log\left(R_{i}A_{k}D_{j}\right)
\end{equation}

RA representations conveys the power responses received from objects around the vehicle in Polar coordinates. A sample of the RA representation, along with the annotation, is shown in Figure~\ref{fig:dataset}.


\section{\uppercase{Proposed Model}}
\label{model}

\begin{figure*}
	\begin{center}
		\includegraphics[width=\linewidth]{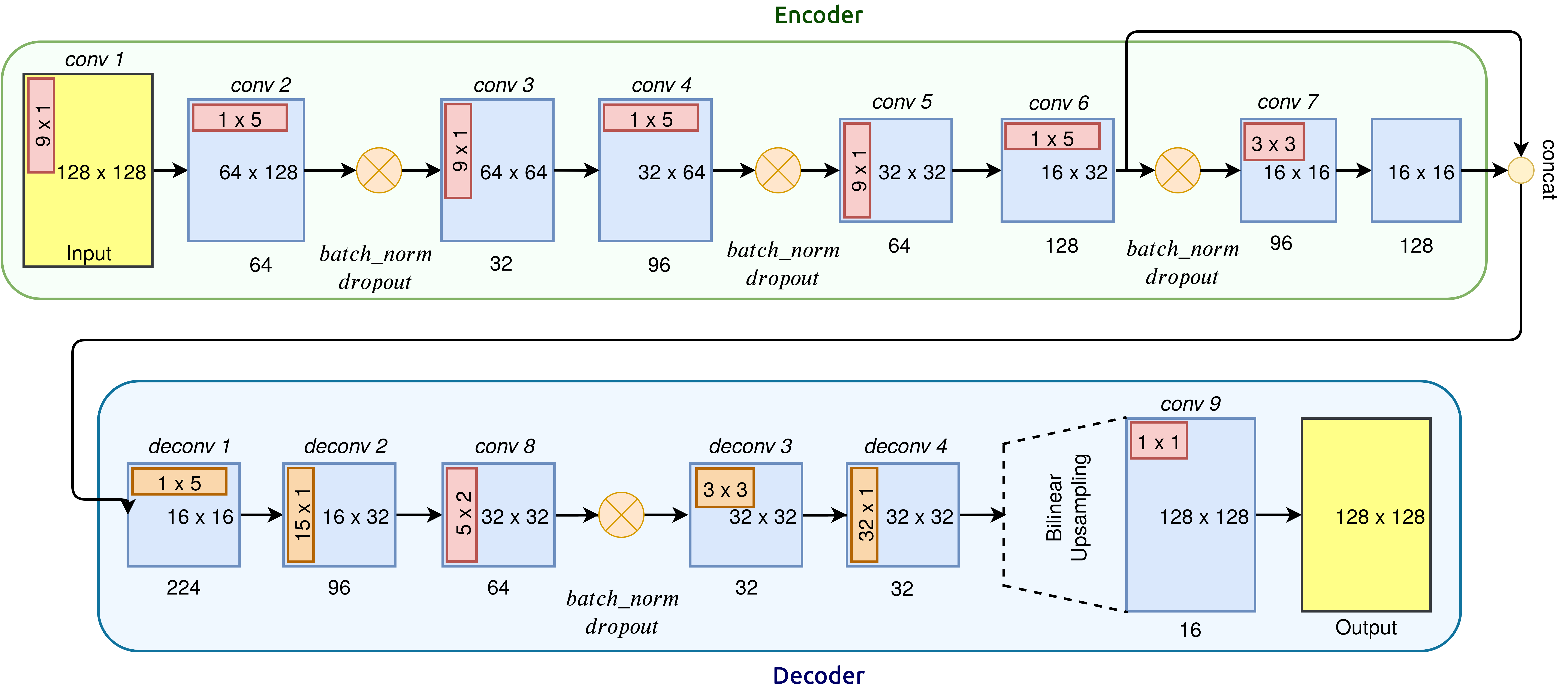}
	\end{center}
	\caption{Detailed architecture of the PolarNet model.}
	\label{fig:polarnet}
\end{figure*}

PolarNet is a convolutional neural network that is applied on the RA and RAD representations of radar observations. In this representation, the first dimension (rows) of the input tensor renders the distance, while the second dimension (columns) corresponds to specific angles. Inspired by StixelNet~\cite{levi2015stixelnet} that uses Stixels, we propose applying one dimensional convolutions on each column of the input. In this way, we apply a filter on each angle and effectively try to identify where the open space ends. Furthermore, by using one-dimensional filters, we are reducing the complexity of the network.

PolarNet is an encoder-decoder network with a smoothing layer at the end of the model. All of the layers use ReLu as the activation, except for the final two layers that employ Sigmoid activations. 

The encoder portion of the model consists of 7 layers. Three of these layers are column-wise convolutional filters that are designed to find the location where the ground meets the first obstacle. Another three layers, positioned after each column-wise layers,  are designed as row-wise convolutions. These layers are used to pool the information from neighbouring columns, and reduce the size and, consequently, the computational complexity of the model. Column-wise layers have a stride of 2 along the columns and row-wise layers use an stride of 2 along each row. After each column-wise and row-wise convolution tuple, a batch-normalization~\cite{ioffe2015batch} and drop-out~\cite{srivastava2014dropout} layer is employed. In the last layer of the encoder, we use a two dimensional $3\! \times\! 3$ convolution with a stride of 1. The output of this layer is concatenated with the output of layer 6 and is then passed to the decoder network.

The decoder network consists of a series of convolutions and transposed convolutions. Reversing the order in the encoder, one row-wise transposed convolution is applied on the feature map, followed by a column-wise transposed convolution. The former layer has a stride of 2 along the columns, while the latter uses the same stride size along the rows dimension. The output of this layer is fed to a two dimensional convolution layer with a kernel size of $5\! \times\! 2$. This layer is responsible for integrating adjacent feature maps with the goal of reducing jittery artifacts in the final output. We apply the final batch-normalization and drop-out layer. Another column-wise convolutional layer with a larger kernel of $32\! \times\! 1$ is used at this stage, thereby forcing a larger receptive field over the feature map. Radar is capable of observing spaces behind various objects such as cars. Using this filter size, the information from an obstacle is propagated to those trailing areas. This helps in assigning opposite bits to the ground and non-ground locations; if the larger receptive fields are not used, the model will be confused by the conflicting nature of the information in the input and the mask requirements.

\begin{figure*}[h]
	\begin{center}
		\includegraphics[width=.9\linewidth]{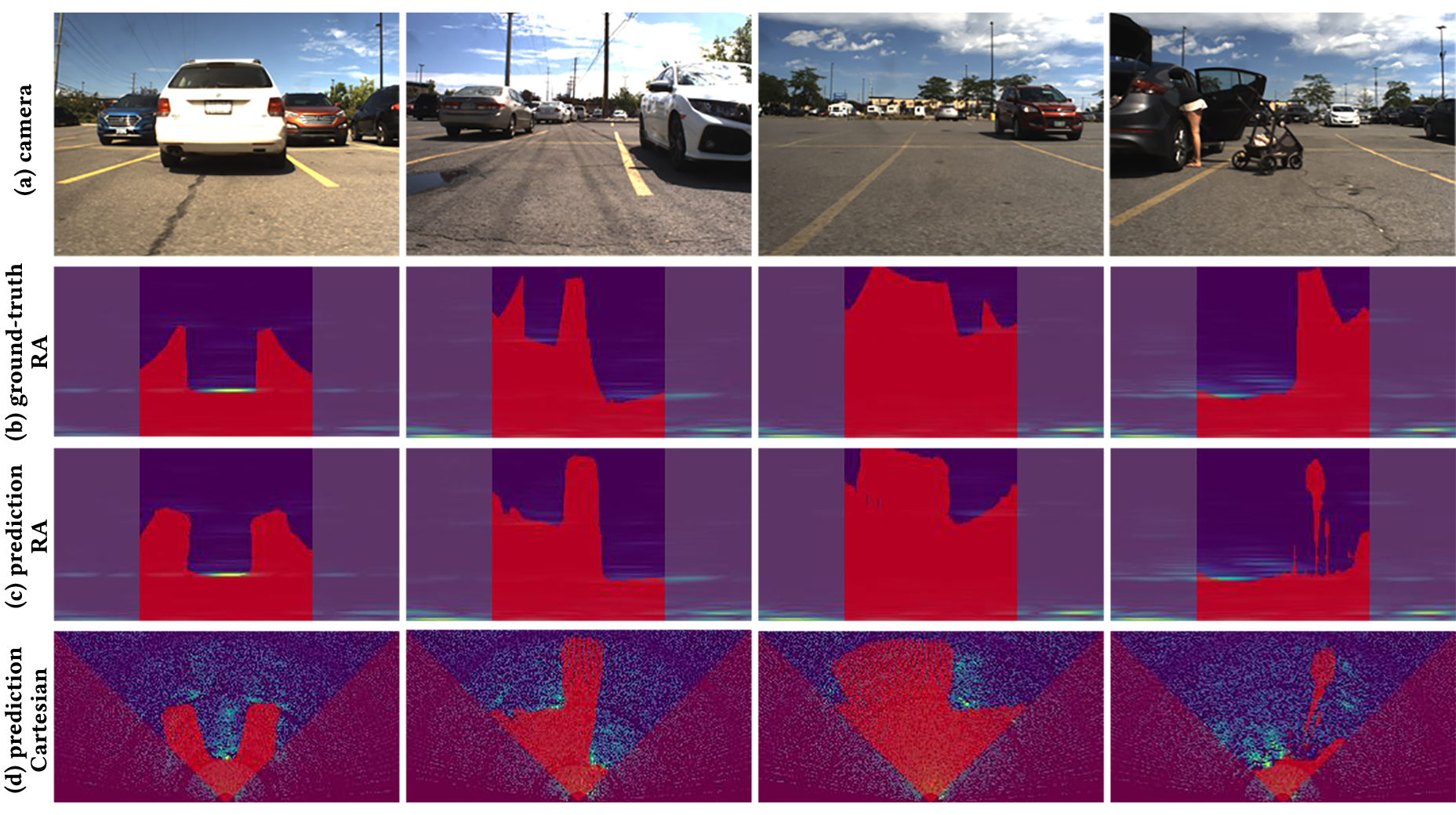}
	\end{center}
	\caption{Sample results of PolarNet. (a) camera image. (b) ground-truth in polar coordinates. (c) segmentation result. (b) segementation result shown in Cartesian coordinates.}
	\label{fig:results}	
\end{figure*}

At each layer of the decoder, the depth of the filter map is consistently reduced. Prior to the final $1\times1$ convolution layer that pools all the information to produce the final result, the feature maps are upsampled to match the input shape through bilinear up-sampling. The complete architecture of the proposed model is shown in Figure~\ref{fig:polarnet}.

The final output contains logits in the range $\left[0,\! 1\right]$. Cells with a value lower than $0.5$ is considered as occupied, while values higher than this threshold are considered as non-occupied. After this division, a softmax cross-entropy (SMCE) function is used to extract the final loss value. We use the $\mathrm{\textit{SMCE}}_{\mathrm{\textit{train}}}$ loss function as in \cite{sc2020icmim}:

\begin{equation}
\resizebox{0.85\linewidth}{!}{$
	\mathrm{\textit{SMCE}}_{\mathrm{\textit{train}}} = \sum\limits_c^{\mathrm{class}} \left( \dfrac{1}{N_c} \sum\limits_{i}^{N}\left( e^{-w_c} \times \mathrm{\textit{SMCE}}_{i}  + |w_c|\right)\right)
	$}
\end{equation}

where $w_c$ represents the trainable weight for class $c$, $N_c$ is the number of pixels of class $c$ in a particular groundtruth mask, and $\mathrm{\textit{SMCE}}_{i}$ represents the softmax cross-entropy loss calculated for a pixel $i$. As such, this modification of SMCE uses trainable parameters to weight the SMCE loss on a per-class basis.

A rate of $0.5$ is used as drop-out probability. Batch size is set at 64. RMSProp with initial learning rate of $0.1$ and decay factor of $0.8$ at every $3500$ steps is chosen as the optimizer. We use Tensorflow \footnote{\href{www.tensorflow.org}{www.tensorflow.org}} as the framework to implement and test our model.


\section{\uppercase{Experiments}}
\label{exps}

We use the radar dataset of \cite{sc2020icmim} to perform our experiments. This dataset consists of $3913$ frames from $11$ sequences with corresponding ground truth. Further, we use their processing pipeline\footnote{\href{www.sensorcortek.ai/publications/}{www.sensorcortek.ai/publications/}} to produce RAD and RA representations to feed as input to our model. 
Two of these sequences are left out to make up the test set enabling us to measure the model's ability to generalize on previously unseen scenarios. This ensures the selection of models that do not overfit on the training set.

The annotated points from the Cartesian ground truth are transformed into the Polar coordinate system to generate the Polar ground truth information. For training, RA and RAD tensors are cropped to match the selected field of view. Note that cropping along columns (angle dimension) equates to reducing the angular field of view in the Cartesian domain.

To benchmark our work we re-use the three deep learning approaches from \cite{sc2020icmim} that include \textit{DeepLabv3+}~\cite{2018_deeplabv3+}, \textit{Fully Convolutional Networks}~(FCN)~\cite{FCN} and thier proposal \textit{FCN\_tiny}. All three of these segmentation models use MobileNet-v2 as their back-bone.

\textit{DeepLabv2} introduced the idea of replacing convolutions with atrous convolutions to increase the effective field of view of a feature extractor without using the feature extractor's last 2 pooling layers. The resulting networks uses much larger feature maps, and thus has a higher memory cost. The latest version of this network, \textit{DeepLabv3+}, uses an atrous spatial pyramid pooling (ASPP) module, introduced in DeepLabv2, on the extractor's output, made up of atrous convolutions of varying rates and an image pooling layer in parallel, to help the network detect objects of varying size. A small decoder, similar to FCN's skip architectures, is also used on the ASPP's outputs upsampled by a factor of four to refine the predictions

As in \cite{sc2020icmim}, we use a complete version of the DeepLabv3+ architecture with rates of 2, 4 and 6 in the ASPP, in addition to a 1x1 convolution and an image pooling layer. We remove the last two pooling layers of MobileNet-v2, and multiply the convolutions' rates by 2 after each removed pooling layer.

The FCN architecture that offers the best compromise between accuracy and speed is the FCN-8's architecture. We use MobileNet-v2 as an encoder and use two skip architectures as a decoder. Each skip architecture works by upsampling its input (either the encoder's output, or the previous skip layer's output) by a factor of two, then concatenating with the feature map of matching size taken from the encoder, which are first passed through a 1x1 convolution to reduce their depth. A 3x3 convolution is applied to the output of the concatenation. As in \cite{sc2020icmim}, we use a depth of 32 within the encoder. After two skip steps, bilinear interpolation is used to resize to the input size.

\cite{sc2020icmim} proposes a smaller version of MobileNet-v2-FCN-8. It uses a depth multiplier of 0.25 in the encoder, and a set depth of 8 in the decoder (25\% of the full FCN's decoder depth).

As in most segmentation literature, we use Mean Intersection-over-Union (Mean-IoU) as the evaluation metric. This metric is calculated as the mean of the IoU's for each class in the dataset. Sample results of the proposed model in both Cartesian and polar forms, along with the matching ground truths and camera images, are shown in Figure~\ref{fig:results}.

Our experiments section is divided into three. First, we compare the effect of various input representations on the performance of the model. Second, we evaluate the effectiveness of various loss functions. Finally, we compare the computational performance and size of our model against the state-of-the-art on multiple platforms.

\subsection{Input Representation}
PolarNet is designed to be applied on polar representations; these include the Range-Azimuth-Doppler (RAD) and the Range-Azimuth (RA) representations. RAD is a tensor of size $128\times128\times64$. Each range bin is approximately $11.17\mathrm{cm}$ with a maximum range of $15\mathrm{m}$ that covers an angle interval of $-45$ to $+45$ degrees. The third dimension represents the Doppler bins that cover a velocity range between $-37.3\, \mathrm{kmph}$ and $+37.3 \, \mathrm{kmph}$. The RA representation is calculated from the RAD representation by selecting the maximum values along the Doppler bins for each Range-Azimuth index.

\begin{table}[h]
	\begin{center}
		\begin{tabular}{|l|c|c|}
			\hline
			Model 		& 	RA input	&	RAD input \\ \hline\hline
			FCN			&	83.98		&	85.16 	  \\ \hline
			DeepLabV3+	&	83.01		&	83.58 	  \\ \hline
			FCN\_tiny	&	84.18		&	83.14 	  \\ \hline			
			PolarNet	&	84.20		&	83.79 	  \\ \hline
		\end{tabular}%
	\end{center}
	\caption{Mean-IoU comparison of models with different inputs.}
	\label{tab:experimental-resullts}
\end{table}

Table \ref{tab:experimental-resullts} shows the comparative results of PolarNet against the state-of-the-art.

Our proposed PolarNet model is the best performer with RA inputs. While using RAD results in a performance reduction for PolarNet, it still takes second place behind the much larger FCN model as shown in Table~\ref{tab:speed-table}. From the experiments, we observe that the larger models have an easier time managing RAD inputs that contain 64 times more information than RA inputs. Inversely, smaller models such as FCN\_tiny and PolarNet have a harder time handling this huge dimensionality increase. As we see in the next sections, the small reduction in performance comes with a larger advantage in the computational complexity of the system.

\subsection{Loss Functions}

We compare three loss functions for this task: \textit{trainable softmax cross-entropy loss}, \textit{softmax cross-entropy} and \textit{Lovasz loss}~\cite{lovaszloss}.

\textit{SMCE} is commonly used as an extension of the typical classification task to the segmentation task; the loss is calculated independently on a per-pixel basis and then averaged over the entire image. To aid in training using SMCE, it is better to use a weighting parameter to reduce effect of any imbalance on data. However, tuning this extra hyper-parameter can be cumbersome. Another interesting function to consider is the \textit{Lovasz} loss. It enables optimization of the mean IoU metric directly, thereby showing a direct link between the main metric for segmentation and the optimization of the loss function. We compare these two functions with the loss function used in our previous experiments: $\mathrm{\textit{SMCE}}_{\mathrm{\textit{train}}}$.

PolarNet architecture is used with RA inputs to evaluated mentioned loss functions. Results are reported in Table \ref{tab:loss-table}.

\begin{table}[h]
	\begin{center}
		\begin{tabular}{|l|l|l|}
			\hline
			Loss Function & Mean-IoU \\ \hline\hline
			$\mathrm{{SMCE}}_{{\textit{train}}}$    & 84.20    \\\hline
			SMCE          & 82.02    \\\hline
			Lovasz        & 82.85    \\ \hline
		\end{tabular}
	\end{center}
	\caption{Mean-IoU for PolarNet with RA input and polar labels using different loss functions.}
	\label{tab:loss-table}
\end{table}

$\mathrm{\textit{SMCE}}_{\mathrm{\textit{train}}}$ achieved the best results in this experiment. Weighting parameter learning in $\mathrm{\textit{SMCE}}_{\mathrm{\textit{train}}}$ results in its better performance in comparison to the original SMCE and Lovasz that both lack the weighting functionality. This has shown to be the key differentiator in the performance of the models.

\subsection{Complexity Analysis}
Our final analysis concerns the computational complexity and the speed of our model architectures. It is crucial for the models to target embedded deployment and real-time performance on such systems. The details of our experiments for the proposed model architectures are shown in Table \ref{tab:speed-table}.

\begin{table*}[h]
	\begin{center}
		\resizebox{\textwidth}{!}{
			\begin{tabular}{|l|c|c|c|c|c|c|c|c|}
				\hline
				Input    & \multicolumn{4}{c|}{RA} & \multicolumn{4}{c|}{RAD}   \\ \hline
				Model    & PolarNet & FCN\_tiny & FCN     & DeepLabv3+ & PolarNet & FCN\_tiny   & FCN    & DeepLabv3+      \\ \hline\hline
				GPU~(fps)     & 575.01     & 324.50    & 300.75     & 275.15    & 364.89     & 249.79       & 224.56    & 199.83  \\\hline
				CPU~(fps)     & 271.39  & 267.58    & 118.56   & 61.91    & 208.14   & 189.12   & 133.78  & 61.93 \\\hline
				TX2 GPU~(fps) & 54.65   & 31.91      & 29.18    & 22.39     & 48.18 & 28.87    & 28.91   & 21.46   \\\hline
				TX2 CPU~(fps) & 25.90   & 41.62   & 22.20   & 10.46   & 17.97    & 36.86   & 19.68  & 10.09    \\\hline
				\# parameters & 562,472  & 210,279  & 2,933,449  & 3,223,865  & 598,758  & 214,817   & 2,951,593   & 3,242,009  \\\hline
				Memory cost   & 29.85~Mb & 16.02~{Mb} & 59.64~Mb & 134.86~Mb & 39.79~Mb & 23.04~Mb & 70.81~Mb & 143.74~Mb \\\hline
			\end{tabular}
		}
	\end{center}
	\caption{Computational complexity comparison for the tested methods.}
	\label{tab:speed-table}
\end{table*}

DeepLabV3+ is the largest model with slowest performance in both RA and RAD input cases. Surprisingly, the segmentation performance of this model is also inferior to the other models. FCN is 90\% of the size of DeepLabV3+ and almost 10\% faster on a \textit{Geforce RTX2080 TI} GPU. However, on a \textit{Core i9-9900K} CPU, it is almost twice as fast. This trend can be observed on the Jetson TX2 as well.

FCN is still a large model when considering a Radar on Chip (RoC) product. PolarNet and FCN\_tiny address this challenge. PolarNet is one fifth of the size of DeepLabV3+. It runs more than $2.5$ times faster on Jetson TX2 while providing the best mean-IoU on RA and second best on RAD.

The main advantage of FCN\_tiny against PolarNet is its smaller parameter space. FCN\_tiny perfoms faster on the Jetson's CPU than PolarNet, but PolarNet outperforms FCN\_tiny on the Jetson TX2 GPU. This is similar to the behaviour of DeepLabv3+ and FCN. In case of PolarNet, this can be attributed to the fact that PolarNet uses larger convolution kernels that are faster to calculate on GPU than on CPU because of their vectorized computation.


\section{\uppercase{Conclusion}}
\label{conclusion}

In this paper, we proposed a novel deep model, PolarNet, to segment open spaces in parking scenarios using automotive radar. Our model takes benefit of vectorized GPU operations and outperforms the state-of-the-art in terms of speed.

Furthermore, PolarNet provides the state-of-the-art performance using Range-Azimuth as its input modality. These characteristics of the model make it the perfect candidate for RoC integration scenarios.

We have proposed a single frame model. To further increase the performance of this model, temporal filtering methods could be used: aggregating predictions at various timesteps would reduce the noise in the segmentation mask. However, this will result in more parameters and larger computational requirements that needs to be addressed prudently.



\bibliographystyle{apalike}
{\small
\bibliography{example}}

\end{document}